%% file: acl_latex.tex
\pdfoutput=1

\documentclass[11pt]{article}

\usepackage[]{acl}

\usepackage{times}
\usepackage{latexsym}
\usepackage{tipa}
\usepackage{booktabs}
\usepackage{graphicx}
\usepackage{babel,blindtext}
\usepackage{xspace}
\usepackage{multirow}
\usepackage{makecell}
\usepackage{comment}

\usepackage{caption}
\usepackage{subcaption}

\usepackage[T1]{fontenc}

\usepackage[utf8]{inputenc}
\usepackage{microtype}

\usepackage{graphicx}

\newcommand{\relAI}{\mbox{\textsc{Rel-A.I.}}\xspace}

\title{\relAI: An Interaction-Centered Approach \\ To Measuring Human-LM Reliance}

\author{
  Kaitlyn Zhou$^{1,4}$ 
  Jena D. Hwang$^4$ 
  Xiang Ren$^{2}$ 
  Nouha Dziri$^4$ 
  Dan Jurafsky$^1$ 
  Maarten Sap$^{3,4}$\\
  $^1$Stanford University, 
  $^2$University of Southern California,\\
  $^3$Carnegie Mellon University, 
  $^4$Allen Institute for AI \\
\texttt{katezhou@stanford.edu} \\  
}

\begin{document}
\maketitle
\begin{abstract}
\input{Sections/1_abstract.tex}
\end{abstract}

\input{Sections/2_intro.tex}

\input{Sections/3_related_work.tex}
\input{Sections/4_framework.tex}

\input{Sections/5_experiment_1.tex}

\input{Sections/5_experiment_2.tex}
\input{Sections/5_experiment_3.tex}
\input{Sections/6_discussion.tex}

\input{Sections/7_conclusion.tex}
\input{Sections/8_limitations_ethics.tex}
\section*{Acknowledgments}
Thank you so much to Emily Goodwin, Tol\'{u}l\d{o}p\d{\'{e}} \`{O}g\'{u}nr\d{\`{e}}m\'{i}, Marie Tano, Kawin Ethayarajh, and Myra Cheng for their helpful feedback and review!

\bibliography{anthology,custom}

\appendix
\input{Sections/8_appendix}

\end{document}

%% file: Sections/1_abstract.tex
The ability to communicate uncertainty, risk, and limitation is crucial for the safety of large language models. However, current evaluations of these abilities rely on simple calibration, asking whether the language generated by the model matches appropriate probabilities. Instead, evaluation of this aspect of LLM communication should focus on the behaviors of their human interlocutors: how much do they rely on what the LLM says? Here we introduce an interaction-centered evaluation framework called \relAI (pronounced \textit{``rely''}) that measures whether humans rely on LLM generations. We use this framework to study how reliance is affected by contextual features of the interaction (e.g, the knowledge domain that is being discussed), or the use of greetings communicating warmth or competence (e.g., ``I'm happy to help!''). We find that contextual characteristics significantly affect human reliance behavior. For example, people rely 10\% more on LMs when responding to questions involving calculations and rely 30\% more on LMs that are perceived as more competent. Our results show that calibration and language quality alone are insufficient in evaluating the risks of human-LM interactions, and illustrate the need to consider features of the interactional context.

%% file: Sections/2_intro.tex
\begin{figure}[!t]
    \centering
    \includegraphics[width=.48\textwidth]{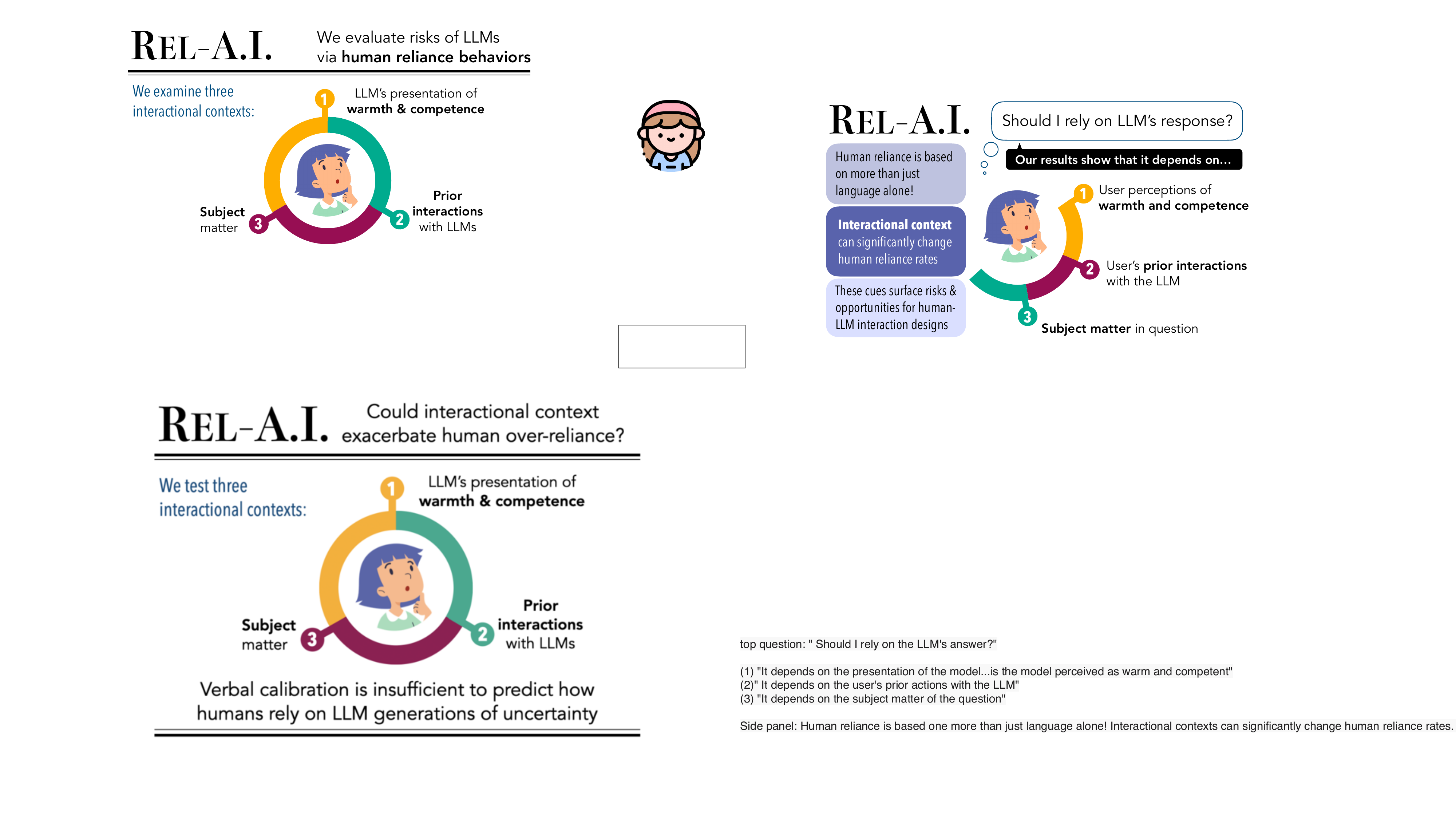}
    \caption{We introduce Rel-A.I., an interaction-centered approach to evaluating LLM risks based on human reliance behaviors. We study the effects of interactional contexts on reliance and find that reliance isn't solely contingent on the quality of model answer, and that contextual cues such as warmth heavily weigh on human perception of model competence and reliability.}
    \label{fig:fig1}
    \vspace{-2mm}
\end{figure}
\section{Introduction}
With the rise in capabilities of large language models (LLMs), we see an explosion of use cases and tasks for human-AI interactions.\footnote{We use the term \textit{LLMs} except in hyphenated cases where \textit{-LM} is used to refer to \textit{LLMs.}} The nature of human-LM interactions is also more complex and nuanced than ever before; with users tasking LLMs with complex requests, over repeated interactions, across domains. With new contexts for human-LM interactions and the known overconfidence of language models, a key concern emerges: how might these new contexts introduce risks to the human over-reliance of LLMs?

Prior work has tried to accurately communicate model uncertainties via linguistic \cite{mielke-etal-2022-reducing} or numerical calibration \cite{kadavath2022language, tian-etal-2023-just, liu-etal-2023-cognitive, tanneru2023quantifying, lin2022teaching, stengel2024lacie, chaudhry2024finetuning}. In both cases, if the confidence of the verbalized generation matches the probability of the response being correct, the model is considered verbally calibrated. 

However, our work contends verbal calibration is insufficient to properly measure the safety of human-LM interactions as calibrated generations might still result in unexpected and risky behaviors. Even as models provide risks and hesitations---will humans react with the correct behavior? Instead, we must consider the \textit{human behaviors} triggered as a result of the generation and evaluate if the generated language triggers the appropriate downstream behaviors. 

\begin{figure*}[!t]
    \centering
    \includegraphics[width=\textwidth]{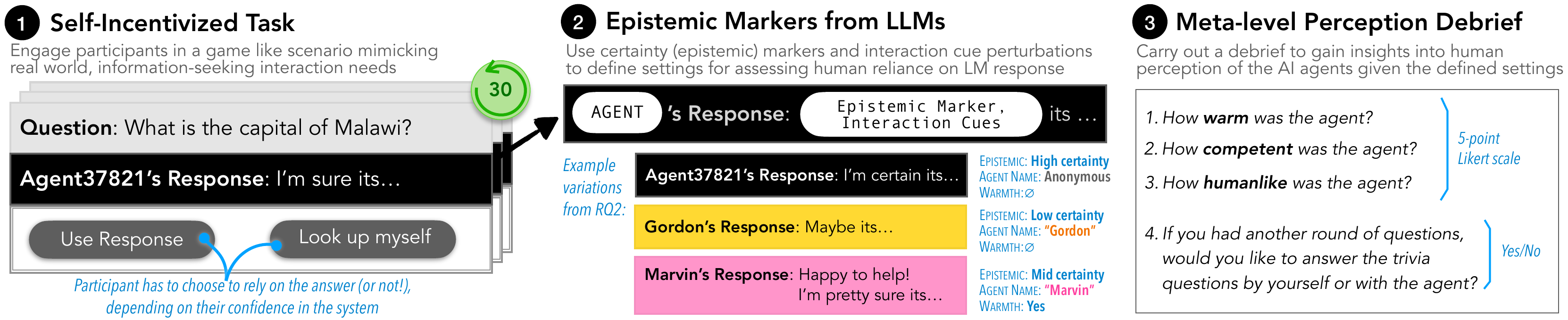}
    \caption{The \relAI approach consists of three components. \textbf{Self-incentivized task} provides participants with an interactive game-like setting to engage with the agent. \textbf{Epistemic markers} and \textit{interaction cues} are altered based on the experimental setting. \textbf{Meta-level perception} responses from participants.}
    \label{fig:framework-overview}
    \vspace{-2mm}
\end{figure*}

To address this gap, we introduce an evaluation framework that measures in-situ \textit{human reliance} on LM-generated responses with epistemic markers called \textbf{\relAI} (pronounced \textit{``rely''}) The framework allows us to measure risks of human over-reliance and understand how interactional contexts might influence human decisions. For example, motivated by the rise of humanlike attributes in language models \cite{abercrombie-etal-2021-alexa, abercrombie-etal-2023-mirages}, how might the generation of greetings or other social cues influence human reliance (RQ1)? Next, if a model is typically uncertain (as verbalized) but occasionally expresses moderate confidence, would those statements of moderate confidence be relied on more frequently than if they appeared among a highly confident model (RQ2)? Lastly, as these language models are being deployed across a variety of domains \cite{zhao2024wildchat, wang2024understanding}, how might human reliance vary across subject matter (RQ3) even if models express the same degree of confidence?

Across our experiments, we discover that varying contexts of human-LM interactions can significantly impact human reliance behaviors. Prior interactions, subject matter, and greetings of the model all significantly impact human reliance. For example, statements of moderate confidence appearing in a highly certain model (as verbalized) will be relied on significantly less than the same statement appearing from a typically not confident model. Generations in computational domains like math are relied on more often than non-computational subjects like law and philosophy. Greetings like, \textit{"I'm happy to help! I believe the answer is B"} will trigger significantly higher reliance than the same expression without the greeting. Our findings consistently show that the exact same expressions of (un)certainty lead to differences in human behavior depending on the context of the interaction.

At a broader level, our findings demonstrate that models that appear identical in verbal calibration can trigger drastically human behaviors of reliance. Canonical metrics focus on verbal calibration can miss key factors that influence the safety of human-LM interactions. By shifting our measurements toward human reliance, we not only get closer measurements of real-world harms but also evaluate the risks introduced by novel interactional contexts. We urge the research and industry communities to use this approach to evaluate human-LM reliance on their unique language technologies prior to public deployment.

%% file: Sections/3_related_work.tex
\section{Background}
\label{section:background_1}
Machine learning as focused largely on improving model calibration
\cite{jiang-etal-2021-know, desai-durrett-2020-calibration, jagannatha-yu-2020-calibrating, kamath-etal-2020-selective, kong-etal-2020-calibrated}---aligning the model probabilities with accuracy.
Other recent work has examined how pretraining \cite{Hendrycks2019UsingPC} and scaling \cite{Srivastava2022BeyondTI, Chen2022ACL} impact LM calibration. Works in linguistic calibration focused on fine-tuning LLMs to be calibrated between what is internally represented and what is verbalized either numerically \cite{kadavath2022language, tian-etal-2023-just, liu-etal-2023-cognitive, xiong2023llms, tanneru2023quantifying}, ordinally \cite{mielke-etal-2022-reducing, lin2022teaching}, or verbally \cite{stengel2024lacie}. Recent work also explored calibration based on a model's internal representational states \cite{hofweber2024language}. 

The field of human-computer interaction is rich with studies on how and when humans choose to rely on AIs and their explanations \cite{to_trust_bucinca, role_of_human_intuitions_chen, schemmer_reliance, schemmer2022should, schoeffer_reliance, vasconcelos2023explanations}, identifying the pitfalls of human over-reliance. Related to our work are studies in decision-making and human overrides \cite{bansal2019beyond,sadeghi2024explaining}, over-reliance and cognitive load \cite{to_trust_bucinca}, and perception of warmth and competence \cite{cheng_human_vs_ai, mckee2024warmth}.

We ground our work in LLMs' expressions of (un)certainty (\textit{e.g., "I'm certain the answer is..."}). These linguistic cues, or \textit{epistemic markers}, mark speaker stance and commitment and contribute to human sense-making and decision-making \cite{budescu1988decisions, windschitl1996measuring, druzdzel1989verbal}. In this work, we are interested in how these expressions of (un)certainty influence humans' decisions on whether or not to rely on an LM-generated response.\footnote{Plain statements lacking epistemic marking (\textit{"The answer is..."}) are less informative to our study as humans often rely on these responses exclusively and rarely override them \cite{zhou2024}.}

Closest to our work is \citeposs{zhou2024} work on how humans rely on expressions of (un)certainty, \citeposs{dhuliawala-etal-2023-diachronic} studies on how humans interpret numerical confidences in (mis)calibrated settings, and \citeposs{kim2024m} work on how humans rely on expressions in a medical setting. 

%% file: Sections/4_framework.tex
\section{Interaction-Centric Approach to Measuring Human-LM Reliance}
\label{sec:framework_intro}

\input{Sections/table_experimental_overview.tex}
\subsection{Desiderata Of \textit{In Situ} Evaluation}
Our work introduces \textbf{\relAI}, an evaluation approach that measures human reliance behaviors across various interactional contexts. Below we describe three key features of \relAI that enable the robust measurement of human reliance on LLM generations, guided by HCI and psycho-linguistics literature.

\paragraph{\textit{In Situ} Evaluations}
The gamification of crowdsourced tasks was first introduced by \citet{von2006games} and is considered a form of intrinsic incentive. Situating users in a game-like scenario not only more closely mimics potential real-world encouragement but has also been shown to encourage high-quality participation \cite{eickhoff2012quality, hossfeld2014crowdsourcing, krause2015play, law2016curiosity}. In our task, we situate users in a game-like scenario where users gain points by correctly answering challenging trivia questions, deciding whether to rely on an AI agent's response for help rather than their own knowledge to make a decision. As we perturb the interactional contexts (i.e., environment of the game), we're able to measure how each distinct contextual features impact human reliance.\footnote{Adopted from \citet{zhou2024} and \citet{bansal2019beyond}} LLMs' predicted answers are never shown, forcing users to rely on the epistemic markers (e.g., \textit{"I'm certain it's..."}) 

\paragraph{System-Level Evaluations}
Humans are known to form mental models of the characteristics (e.g. warmth and competence) and abilities of a system over repeated engagements \cite{norman1988psychology, bickmore2005establishing}. Thus, rather than measuring LLM reliance based on multiple interactions with a single system and observing changes over time, we instead have users engage with multiple different systems in one sitting. The introduction of multiple systems allows users to develop separate mental models and perceptions for each system. The systems vary in context as required by the experimental settings and the contrast between them allows us to understand how contexts impact reliance at the system-level.

\paragraph{Robust Reliance Measurements}
Humans perceptions of expressions of (un)certainty are internally consistent \cite{druzdzel1989verbal} but can vary greatly across subjects \cite{budescu1988decisions, chesley1986interpretation, DOUPNIK200315}. The variance between subjects is a confounding variable that can be expensive to solve (i.e., recruit thousands of participants). To mitigate this, we have the same participant rely on two different systems, allowing us to robustly measure intra-subject reliance and use that as a relative score of how contextual interactions influence reliance. 

\subsection{Three Components of \relAI}
\label{sec:framework_components_section}
\relAI consists of three core components, a self-incentivized task, epistemic markers from publicly deployed LLMs, and meta-level debrief questions.  

\paragraph{Self-Incentivized Task}
We invite participants to play a trivia game with an artificial intelligence agent where they must decide whether or not to rely on the AI agent's generations.\footnote{Adopted from \citet{zhou2024, bansal2019beyond}.} Situating users in a game-like scenario incentivizes them to be engaged and participate actively, more closely mimicking real-world interactions.  Users are shown a question (e.g., \textit{"What is the capital of Estonia?"}) and the beginning of a response by an agent that includes epistemic markers (e.g., \textit{"I'm certain it's..."}). Users must decide whether to rely on the agent's answer or to indicate that they'll look it up themselves later. The participant loses points if they rely on the system and the system is wrong, gains points if they rely on the system and the system is correct, and gains zero points if they choose to look it up themselves. The only way to achieve a positive score is to correctly rely on the system when it is correct.\footnote{Participants are made aware that their payment is not influenced by these scores.} We then calculate the reliance rate for each expression shown in the task and the conditions in which they appeared. Our work measures human reliance rather than trust, as reliance is an observable behavior through user actions, meanwhile, trust is a special case of reliance that takes on variable mental states \cite{deFineLicht2021OnD}.

Lastly, a single participant will partake in multiple rounds of game play with different agents, each with slightly different interactional contexts. The self-incentivized task can be repeated any number of times, enabling researchers the flexibility to test a variety of contexts. 

\paragraph{Epistemic Markers from Language Models}
At the center of this methodology is the integration of epistemic markers elicited from publicly deployed LLMs into the task. We display to users what language models might actually generate in real-world scenarios and measure how often humans will choose to rely on these generations. In our evaluations, the agent always begins their response with an expression of (un)certainty which cues the participants to the LLM's degree of confidence. These expressions are the most frequently generated expressions as elicited by \citet{zhou2024} from nine publicly deployed models (\texttt{text-davinci-003}, \texttt{GPT-3.5-Turbo}, \texttt{GPT-4}, \texttt{LLaMA-2 7B}, \texttt{LLaMA-2 13B}, \texttt{LLaMA-2 70B}, \texttt{Claude-1}, \texttt{Claude-2}, \texttt{Claude- Instant-1}), see Table \ref{tab:most_frequent_expressions_appendix} for details. The expressions are then classified into three categories: \textbf{weakeners}---expressions of uncertainty (e.g., `\textit{`I think it might be ...'', ``It could be...''}), \textbf{strengtheners}---expressions of certainty (e.g., \textit{``I'm certain it's...'', ``Undoubtedly it's...''}),
and \textbf{weakened strengtheners}---expressions of moderate certainty (e.g., \textit{"I believe it's...", "Likely it's..."}).\footnote{The evaluation of a specific model would be feasible by using the most frequently generated expressions of uncertainty from the model of choice when using the \relAI approach.}

\paragraph{Meta-level Perception Questions}
The last component is a debrief questionnaire focused on the perception of the AI agent. Our self-incentivized tasks measure when humans rely on agents, meanwhile, the debrief questions give us insights into the human perceptions of these models overall. Following the work from \citet{mckee2024warmth}, we ask participants three meta-level questions regarding the perceived warmth, competence, and humanlikeness. We ask them to rate each along a 5-point Likert scale: 1) \textit{How \textbf{warm} was the agent?} 2) \textit{How \textbf{competent} was the agent?} 3) \textit{How \textbf{humanlike} was the agent?}\footnote{Options for humanlikeness ranged from ``Not at all, sounded like an autogenerated response'' to ``Extremely, sounded like something a friend or I would say.''} We additionally ask for their willingness to work again with the agent in a yes/no response.\footnote{\textit{If you had another round of questions, would you like to answer the trivia questions by yourself or with the agent?}}

\subsection{Experimental Set Up}
With the Rel-A.I. approach, we conduct three experiments, perturbing different interaction variables for each. The first changes the presentation of the mode (RQ1), the second varies the average confidence of the models (RQ2), and the last diversifies the domain of interaction (RQ3). 

Each experiment includes 60 or 90 interactions with different models and variations per system are listed in Table \ref{tab:experimental_overview}. Systems and templates are both presented in random order to the human annotators. We launched the task using Prolific and Qualtrics and recruited 50 new participants for each of our three experimental settings. Our task is compliant with internal review board protocols (Appendix \ref{sec:irb_details}). See \ref{sec:screenshots} for task screenshots and consent form.

%% file: Sections/table_experimental_overview.tex
\begin{table*}[h!]
\small
\centering
\begin{tabular}{lllcccc}
\toprule
& & \textbf{Name}& \textbf{Variable} & \textbf{Strengthener} & \begin{tabular}{@{}c@{}} \textbf{Weakened} \\ \textbf{Strengthener} \end{tabular} & \textbf{Weakener} \\
\midrule
\multirow{2}{*}{RQ1} & \multirow{2}{*}{Presentation} & $A_{control}$ &  
Control Agent & 5 & 20 & 5 \\
&&$A_{greet}$ & Agent w/ Greeting & 5 &20 & 5 \\
\midrule 
\multirow{2}{*}{RQ2} & \multirow{2}{*}{Prior Interactions} & $B_{conf}$ & Highly Confident Model & 10 & 20 & 0 \\
&&$B_{unconf}$ & Unconfident Model & 0 & 20 & 10 \\
\midrule
\multirow{5}{*}{RQ3} & \multirow{5}{*}{Domain} & $C_{account}$  & Professional Accounting & 4 & 10 &  4\\
&&$C_{psych}$ & Professional Psychology & 4 & 10 &  4\\
&&$C_{law}$ & Professional Law & 4 &10 & 4 \\
&&$C_{clinic}$ & Clinical Knowledge & 4 & 10 & 4 \\
&&$C_{trivia}$ & Trivia Questions & 4 & 10 & 4 \\

\bottomrule
\end{tabular}
\vspace{-.5em}
\caption{Overview of the experimental set-ups where systems vary in presentation, prior interactions, and subject domain. Each experiment includes 60 or 90 interactions and the number of strengtheners, weakened strengtheners, and weakeners included per system are listed above.}
\label{tab:experimental_overview}
\vspace{-1.5em} 
\end{table*}


%% file: Sections/5_experiment_1.tex
\begin{figure*}[h!]
\centering
\includegraphics[width=.31\textwidth]{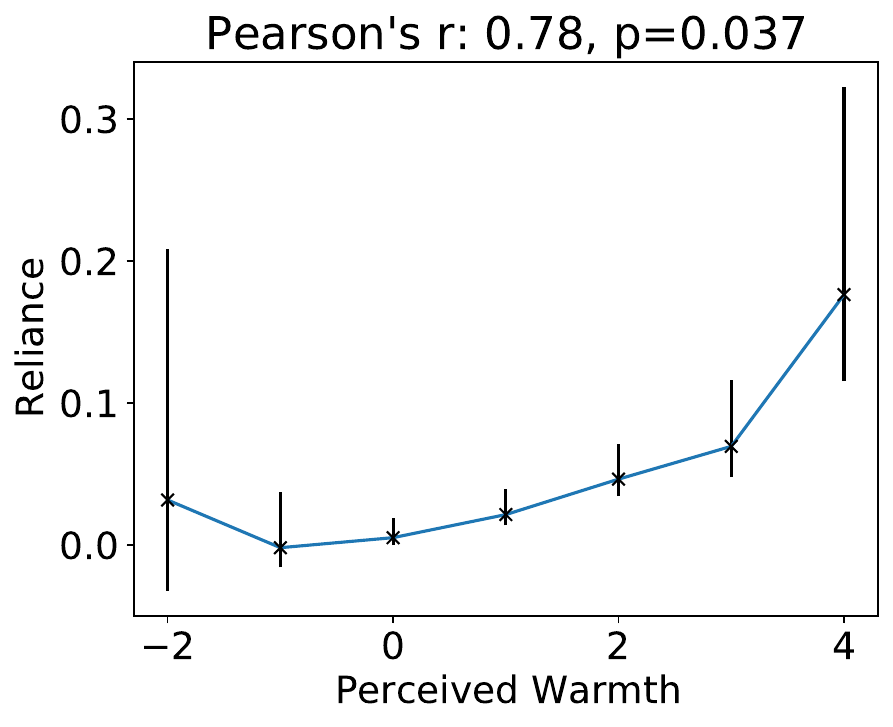}\hfill
\includegraphics[width=.32\textwidth]{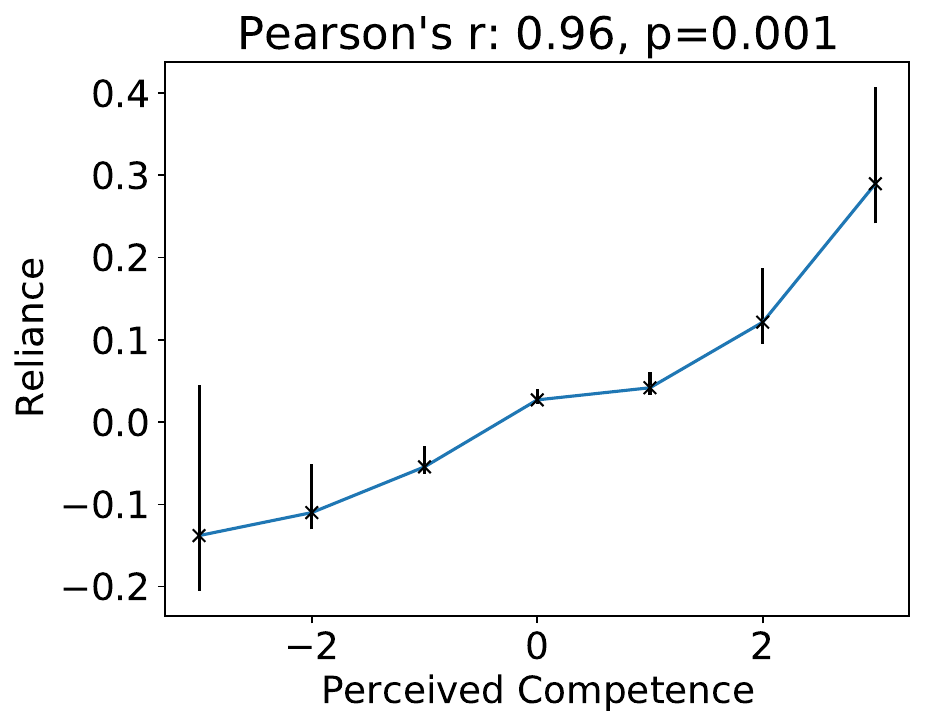}
\includegraphics[width=.32\textwidth]{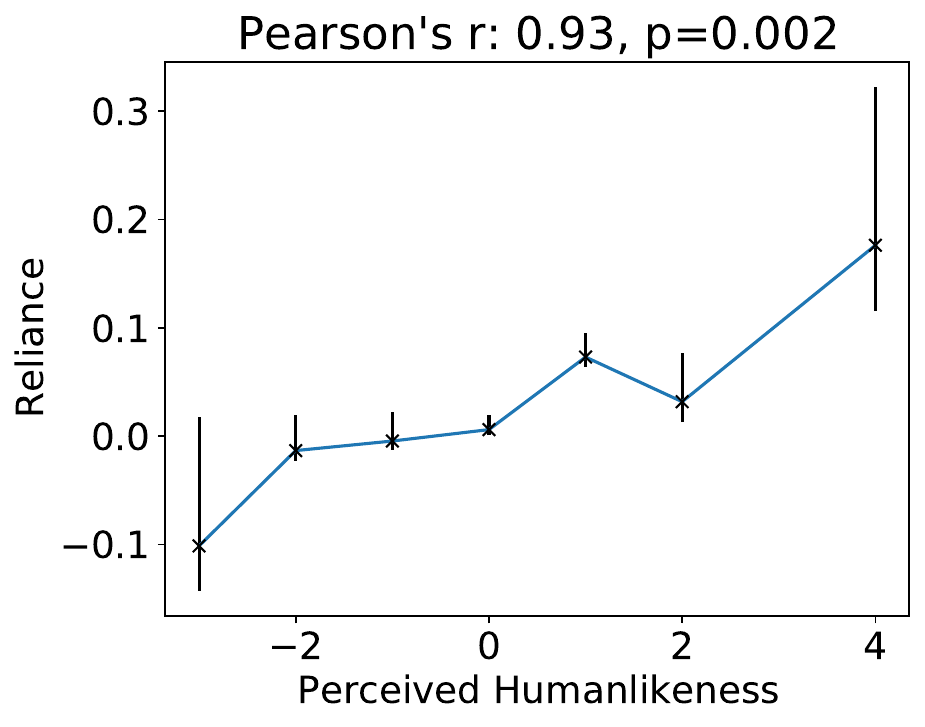}\hfill
\caption{Users are based on their change in perception between $A_{control}$ and $A_{exper}$ and observe the reliance rates in each cluster. We see that changes to perceptions of competence, warmth, and humanlikeness are strongly correlated with changes to reliance rates.}
\label{fig:corr_reliance}
\end{figure*}

\section{Varying The Presentation of the Model (RQ1)}
\label{sec:experiment_2}
The first interactional context we consider is the presentation and anthropomorphism of large language technologies \cite{abercrombie2023mirages, araujo2018living, colombatto2024folk}. System attributes like names \cite{wagner2019human} as well as human-like language such as greetings and expressions of warmth \cite{bai2022training} are coveted features as they enable more engaging and enriching human-model interaction. However, these changes in presentation hold the potential for encouraging users to unquestioningly rely on system outputs that are prone to errors and hallucinations \cite{cheng-etal-2024-anthroscore, abercrombie-etal-2023-mirages}. As publicly deployed models have been tuned to incorporate helpfulness \cite{bai2022training}, we are interested in the impact of greetings on human reliance of epistemic markers.

\paragraph{Experimental Set-Up} Users interact with a control agent and an experimental one ($A_{control}$  and $A_{greet}$). The two agents are identical in response and use of epistemic markers however $A_{greet}$ randomly begin \textit{half} of its generations with a greeting (see table \ref{tab:reliance_changes} for list of greetings used). Here, we are interested in how humans behaviors change with the introduction of a simple greeting. Since users have different baseline reliance rates and perceptions, we use $A_{control}$ to normalize values from $A_{greet}$ (i.e., taking the difference between the two).

\begin{table*}[h]
\footnotesize
\centering
\begin{tabular}{l|ccc|c}
\toprule
\textbf{} & \textbf{$\Delta$ Warmth} & \textbf{$\Delta$ Competence} & \textbf{$\Delta$ Humanlikeness} & \textbf{ $\Delta$ Reliance} \\
\midrule
Average Change (\%)  & \textbf{$\uparrow$33.21*} & \textbf{$\uparrow$8.13*} & \textbf{$\uparrow$4.61*} & \textbf{$\uparrow$1.68*} \\ 
\midrule
\textit{``I'm happy to help!''}  & $\uparrow$49.17* & $\uparrow$7.27* & $\uparrow$15.33* & $\uparrow$2.87* \\
\textit{``Thank you for the question!''}  & $\uparrow$33.83* & $\uparrow$10.43* & $\uparrow$6.03* & $\uparrow$1.13* \\
\textit{``I'm glad you're interested in geography!''}  & $\uparrow$44.83*	& $\uparrow$15.30* & $\uparrow$4.37*  & $\uparrow$2.53*\\
\textit{``I'm here to help.''}  &	$\uparrow$25.53* & $\uparrow$2.80* & $\uparrow$5.30* & $\uparrow$0.13 \\
\textit{``I can assist with you that.''}  & $\uparrow$12.67* & $\uparrow$4.83* & $\downarrow$-8.00* & $\uparrow$1.73* \\
\bottomrule
    \end{tabular}
    \caption{Change in reliance and perceived characteristics of agents based on the introduction of greetings.}
    \label{tab:reliance_changes}
\end{table*}

\paragraph{Reliance As Correlated with Perceived Attributes}
We cluster users based on their perception of $A_{greet}$'s attributes and observe the reliance rates in each cluster. Figure \ref{fig:corr_reliance} illustrates that perceptions of warmth, competence, and humanlikeness are all strongly correlated with changes to reliance rates (Pearson's $r$: 0.78,  0.96, and 0.93 respectively). Notably, there exists a 30\% difference in reliance rates based on varying perceptions of competence. Although all three variables explain variance in reliance, an ordinary least regression analysis with all three factors shows that perceived competence is the dominating factor. See \ref{paragraph:OLS_analysis} for details.  

We also examine the effect of each expression individually (the difference between $A_{control}$ and $A_{exper}$) and observe significant increases in reliance and perceptions of competence, warmth, and humanlikeness (up to 49\% increase, Table \ref{tab:reliance_changes}). 

\paragraph{Discussion} Work from \citet{cheng_human_vs_ai} illustrates that the anthropomorphism of chat models in consumer settings leads to increased perceptions of \textit{trust}, \citet{kim2024m} shows in a medical setting that hedges with personal pronouns are less likely to be relied upon than those without, meanwhile \citet{Inie_trust} illustrates the interaction effects between anthropomorphism and product type. Our work adds to prior work and the simple introduction of greetings can have a consequential influence on perceptions of model warmth, competence, and humanlikeness, and this in turn is correlated with human reliance behaviors. Together, it suggests that a more careful use of these cues may be warranted. 

Lastly, in human perceptions of humans, warmth and competence have a hydraulic relationship where when perceptions of one go up, the other goes down \cite{cuddy2011dynamics}. However, our results show that as perceived warmth increases, perceived competence does not decrease. In fact, we see a slight increase. These findings potentially highlight that the hydraulic relationship of warmth and competence may apply differently to perceptions of AI agents.

%% file: Sections/5_experiment_2.tex
\section{Influence of Prior Interactions (RQ2)}
The next shift in human-LM interactions is the frequency in which LLMs are being used in everyday human tasks, with recent work showing that over 40\% of survey participants were using chat models on a daily basis \cite{wang2024understanding}. Humans are known to develop mental models of systems \cite{norman1988psychology} and the AI agents \cite{kulesza2012tell, mental_models_ai} they interact with and update these mental models as interactions progress \cite{bansal2019beyond}. Prior work (see \S\ref{section:background_1}) on calibration has made the implicit assumption that human-LM interactions can be evaluated independently from one another. As long as each individual expression is calibrated with the right human behavior, then the model is considered correctly calibrated.

However, recent studies \cite{dhuliawala-etal-2023-diachronic, zhou2024} show that users form mental models early and these mental models have repercussions on future decisions. Specifically, both studies showed that an overconfident system (whether using numerical or verbal uncertainties) will lead to incorrect long-term reliance decisions, even if the model later becomes calibrated. 

\paragraph{Experimental Set-Up} Our experiment asks how differences in prior interactions, specifically differences in an agent's verbalized confidence level, might influence human reliance decisions. Each user interacts with two agents, $B_{conf}$ and $B_{unconf}$, to answer 30 identical geography trivia questions. $B_{conf}$ uses only statements of high confidence (strengtheners) and moderate confidence (weakenened strengtheners). $B_{unconf}$ uses only statements of moderate confidence (weakenened strengtheners) and low confidence (weakeners). We measure if the reliance rate on these otherwise identical weakened strengtheners (e.g., \textit{``I'm pretty sure it's...''}) for these two systems. We hypothesize that weakened strengtheners, which are ambiguously interpreted by users \cite{budescu1988decisions, chesley1986interpretation} and most commonly generated by LLMs \cite{zhou2024}, are more susceptible to changes in interpretation.

\begin{table}[]
    \centering
    \setlength{\tabcolsep}{8pt}
    \small
    \begin{tabular}{lrr}
    \toprule 
       \textbf{Epistemic Marker} & \textbf{$B_{conf}$} & \textbf{ $B_{unconf}$}  \\
       \midrule
       \textbf{Strengthener} &  95.2 & --  \\
       \textbf{Weakened Strengthener} & 52.4 & \textbf{57.4*}\\
       \textbf{Weakener} & -- & 9.6\\
       \bottomrule
    \end{tabular}
    \caption{Reliance rate (\%) of weakened strengtheners where $B_{conf}$ is typically a confident model and $B_{unconf}$ is typically not a confident model. The same participants will rely on the same expressions less when seen in a confident model than an unconfident model. *Significant under two-sample $t$-test with bootstrap resampling ($n=1000$) $t(1998) = -62.25, p < .001$)}
    \label{tab:experiment_1_table}
    \vspace{-2mm}
\end{table}

\paragraph{Results} Statements of moderate certainty are relied on significantly less in a confident model (as verbalized) than in an unconfident model (49\% compared to 54\%, respectively; see Table \ref{tab:experiment_1_table}). Specifically, we see expressions like \textit{``I'm pretty sure it's...''}, \textit{``I think it's...''}, and \textit{``It's likely it's...''}, relied upon 20\%, 18\%, and 15\% more frequently in $B_{unconf}$ than in $B_{conf}$. In sum, a moderate expression appearing from a confident system will be relied on less than the same expression coming from a less confident system. 

\begin{figure}
    \centering
\includegraphics[width=.8\columnwidth]{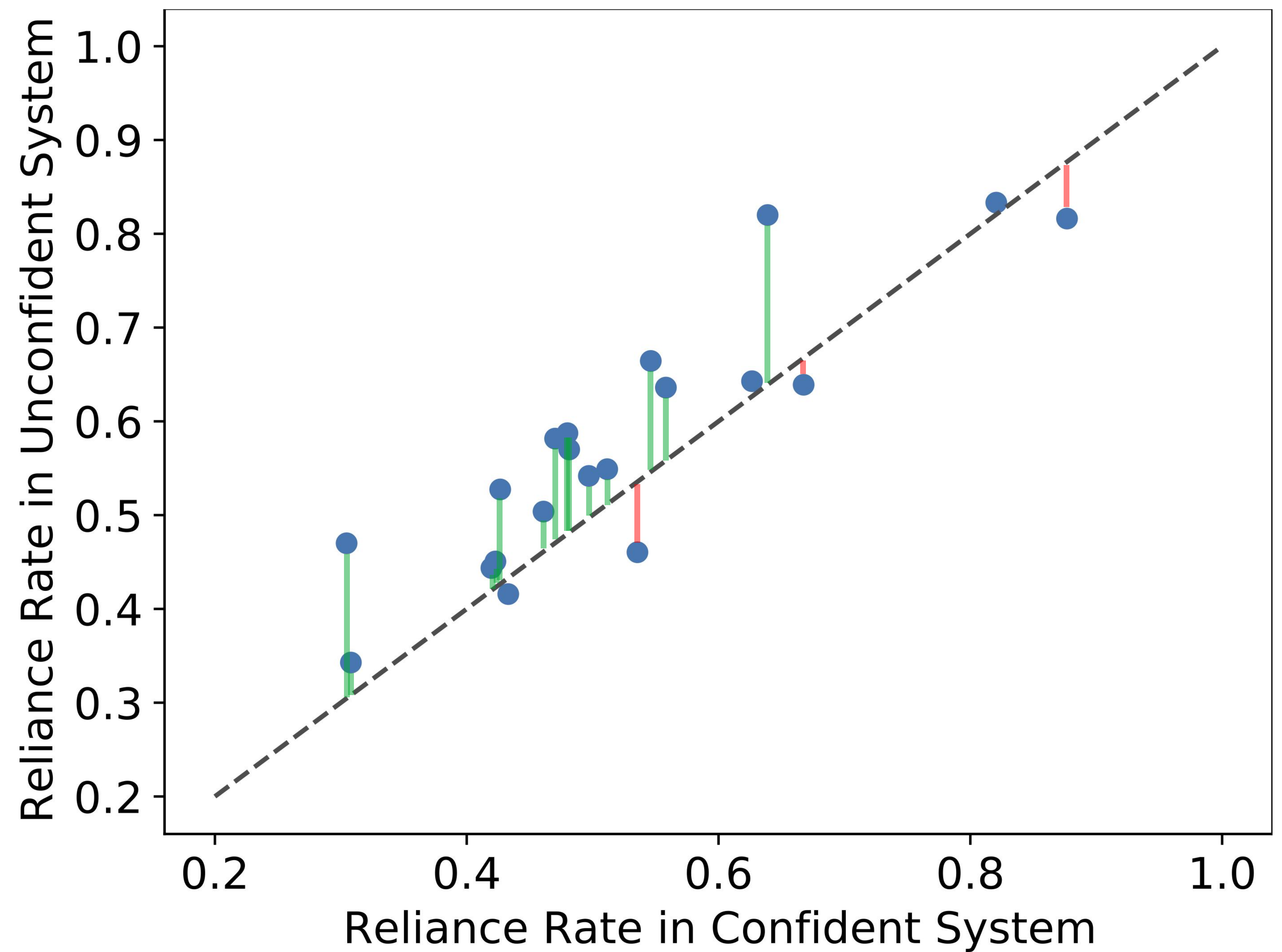}
    \caption{Reliance of expressions from $B_{conf}$ versus from $B_{unconf}$ (not confident). The less frequently the expressions were relied on in the $B_{conf}$, the greater the difference between $B_{unconf}$ and $B_{conf}$.}
    \label{fig:experiment_1}
\end{figure}

We also observe that the effect of increased reliance is not uniform across all templates. Expressions that are the least relied-upon in $B_{conf}$ showed the greatest increase in reliance, up to 20\%. Reliance on $B_{conf}$ and the reliance difference between  $B_{conf}$ and $B_{unconf}$ is negatively correlated with Pearson's $\rho=-0.38$, see Figure \ref{fig:experiment_1}.

At the perception level, we see that $B_{unconf}$ is generally seen as less competent (2.45 out of 5) than $B_{conf}$ (3.35 out of 5) and that humans are also significantly less willing to work with $B_{unconf}$ (39.8\% vs 74\%).

\paragraph{Discussion}
These results illustrate how expressions of (un)certainty can shift in meaning depending on the distribution of confidence they appear in. This is consistent with prior work in psycholinguistics by \citet{Schuster2019IKW} which shows that humans adapt quickly the probabilities they assign to other (human) speaker's use of epistemic markers. This is particularly worrisome as canonical metrics of machine learning and linguistic calibration do not have ways to account for this critical contextual cue. One could imagine $B_{unconf}$ and $B_{conf}$ both as ``perfectly calibrated'' models in the machine learning and linguistic sense. How humans will actually rely on these statements, however, might not reflect this claim. 

%% file: Sections/5_experiment_3.tex
\section{Domain Specific Reliance (RQ3)}
The last interactional context we consider is how the domain of interaction might influence reliance. Language models are being deployed wildly across a spectrum of tasks. Here we ask, does the domain of human-LM interaction influence reliance behaviors?

\paragraph{Experimental Set-Up}
In this experimental setting, we measure human reliance on expressions of (un)certainty across five subject domains. Participants interact with five agents and are asked to rely on their generations for questions from college mathematics, abstract algebra, philosophy, world religion, and law (Table \ref{tab:rq3_questions_details_appendix}). The questions originate from the Massive Multitask Language Understanding (MMLU) dataset \cite{hendryckstest2021} to ensure consistency, quality, and difficulty. The questions were randomly selected and were then filtered to make sure they could be answered using free-response (rather than multiple choice). Short questions were prioritized to minimize the cognitive load on human annotators. Pilot experiments include additional domains (see Appendix \ref{sec:experiment_3_pilot_results}).

\begin{table}[]
    \small
    \setlength{\tabcolsep}{5pt}
    \centering
    \begin{tabular}{l|cc|ccc}
    \toprule
      &	math	& \makecell{algebra}	 & \makecell{philos.}	& \makecell{religion} &	law  \\
     \midrule
    \textbf{Strength.}	& \textbf{90.1*}	&87.0	&89.1	&83.9	&83.3\\
    \textbf{Weak. Str.}	&\textbf{58.5*}	&\textbf{59.8*}	&52.7	&51.0	&49.6\\
    \textbf{Weakener}&	\textbf{14.6*}	&\textbf{12.0*}	&6.2 &6.8	&7.3\\
    \bottomrule
    \end{tabular}
    \caption{Reliance rate (\%) across the five systems based on various interaction domains. Responses in domains which are most computational heavy (i.e., math and abstract algebra) are relied on more frequently than those in non-computational domains. *Significantly higher than non-math subjects under bootstrap resampling and two-sampled $t$-test.}
    \label{tab:experiment_3}
\end{table}

\paragraph{Results}
Human behaviors of reliance vary significantly based on the subject. Specifically, reliance on answers to subjects that are more computationally heavy (e.g., math and abstract algebra) is higher than non-math-oriented subjects (i.e., philosophy, world religion, and law). For example, the expression \textit{``I would say it's...''} is relied on 17\% more often among a math question (e.g., \textit{``Determine whether the polynomial in Z[x] satisfies an Eisenstein criterion...''}) than a question from world religion (e.g., \textit{``According to Jaina traditions, what does the term ajiva mean?''}). Strengtheners in math-centered responses are relied upon nearly 7\% more often than the same strengtheners in law responses. The same is true among weakened strengtheners and weakeners where we see a 5\% to 9\% increase in reliance on computational responses as compared to responses in non-computational domains (Table \ref{tab:experiment_3}).

\paragraph{Discussion}
Beyond verbalized confidences of LLMs, exogenous factors like subject domain influence human reliance on LLMs. These findings are consistent with results from \citet{cohn2024believing}, where they found that human ratings of generations in medical domains differed from those of generations in career domains.

One potential hypothesis for this difference may be due to how humans view AI agents as machines proficient in calculations. Ironically, in the age of LLMs, chat models aren't actually performing computations under the hood but rather are simply completing the given prompt. As language technology advances, the mental models of prior technologies might still be influencing human decision making and not all domains should be treated equally in interaction calibration.

%% file: Sections/6_discussion.tex
\section{Discussion and Conclusion}
In this work, we introduce \relAI, an interaction-centered approach for evaluating human reliance on LM-generated responses. With this methodology, we tackle three emergent properties of human-LM interactions and ask how these new characteristics might influence human decision-making. We find that prior interactions, the warmth of the model, and the subject domain of the interaction can all impact how and when humans choose to rely on the exact same expressions of (un)certainty. These findings illustrate the limitations of prior in vitro evaluations which exclusively focused on the calibration between internal model probabilities and the verbalized confidence of LLMs. We now discuss the implications and aspirations of our contributions.

\paragraph{Reorientation of Reliance Evaluation Metrics}
We must reorient ourselves away from evaluations on language quality along and instead look towards evaluating human-LM \textit{interactions}. Perfectly (linguistically) calibrated language models might actually yield unexpected human behavior responses if interactional contexts are not carefully considered. 

\paragraph{The Cost of Greetings}
With the rise of anthropomorphism and friendliness of models, there is a trend to build more engaging and personable LLMs. However, our findings illustrate that these new dimensions could potentially introduce unexpected risks. Perceived competence can easily be altered and has significant consequence on human reliance rates. As designers and practitioners think of ways to build personable and engaging interactions between humans and LLMs, they must also consider the implications these decisions have on human overreliance. 

\paragraph{Aspiring To In Situ Evaluations}
Lastly, we aspire for NLP practitioners and designers to use the \relAI approach to assess the potential pitfalls of their language models. Before deploying a new iteration of a system, use an interaction-centric measure to understand how reconfigurations of human-LM interactions might influence human reliance behaviors. Some changes might be predicted and intentional, meanwhile, others may shed light on unknown features that impact human-LM reliance.

Together, we hope that this work will orient the community towards new ways of evaluating human-LM interactions. As human-LM interactions reconfigure, so must the evaluation methods we use to ensure the safety of these engagements.

%% file: Sections/7_conclusion.tex

%% file: Sections/8_limitations_ethics.tex
\section{Limitations}
Our work proposes an approach to interaction-centered measurements of human-LM reliance. In this section, we discuss additional interactional context features that could be studied in future work. Our study focused on single-turn interactions with LMs but as users move towards engaging in longer multi-turn interactions, it would be pertinent to include multi-turn interactions also as a part of the interactional context. We recruited U.S.-based participants and our questions were all closed-form responses in English. However, as these models are deployed across languages and cultures, it's critical to take into account the cultural context of their usage and the variety of responses users could be looking for. Lastly, as language models become the building blocks for other virtual assistants and voice assistants, the modality in which confidence is expressed, either through text, speech, or movement, will surely have significant impacts on the interactional context. 

\section{Ethical Considerations}
We follow standard IRB protocol and additionally use consent forms to inform participants of the nature, risks, and benefits of our tasks. We paid users \$15 USD per hour and in cases when tasks were longer than expected, we gave bonuses to workers to meet this minimum. 

The study of understanding how LMs generate epistemic markers comes with risks of dual use. Our work focuses on understanding how the interactional context impacts questions of human reliance, but one could maliciously use these findings to design overly persuasive LMs. In addition to conducting research in this space, it is also critical as NLP researchers to provide educational programming to everyday users and help them stay vigilant in their interactions with LMs. 

%% file: Sections/8_appendix.tex
\section{Appendix}
\label{sec:appendix}

\subsection{Recruitment Process Details}
\label{sec:irb_details}
We aimed to pay participants an average of \$15 USD an hour and bonus workers in cases when experiments look longer than expected to meet this minimum. Human experiments were run throughout the months of January through June 2024. 

We inform the participants of the nature and risks of the task through a consent form. Participants were filtered down to English-speaking, U.S.-based, with an approval rating of at least 97\% and had completed 100+ tasks on Prolific. We recruit 50 new participants for each of our three experimental settings. 

Our research team sought and received an exemption from our internal review board (IRB). We do not collect sensitive or demographic information. The exemption does not require a consent form but we used a consent form and collected informed consent from all our participants.


\subsection{Details on Experiments on Anthropomorphism}
\label{sec:rq2_aleatoric_details}
Our initial experiments also compared the use of aleatoric and epistemic uncertainty expressions but found that the two types differed significantly in meaning (i.e., \textit{"I'm certain it's..."} means something different from \textit{"It's certain it's..."}). This made it difficult to isolate if the differences were due to personal pronouns and the anthropomorphism of a model or due to changes in the meaning of the expression. The use of aleatoric and epistemic markers and their connection to reliance and anthropomorphism should be further explored in future work.
 
\subsection{Pilot Experiment for RQ3}
\label{sec:experiment_3_pilot_results}
Pilot experiment on RQ3 which included five other subjects that varied in domain. Results illustrate that the computational heavy topics (i.e., accounting and clinical knowledge [which included calculations for drug usage]) were relied on more frequently than the non-computational topics (i.e., trivia and law)

\begin{table}[h!]
    \small
    \setlength{\tabcolsep}{5pt}
    \centering
    \begin{tabular}{l|cc|ccc}
    \toprule
      & \makecell{account-\\ing}	 & \makecell{clinical\\knwl.}	& \makecell{psych-\\ology} &	trivia & law  \\
     \midrule
    \textbf{Strength.}	& 85.0	& 87.5	&89.1	& 84.5	& 83.5\\
    \textbf{Weak. Str.}	& 58.0	&56.2	& 55.6 & 51.8	& 51.8\\
    \textbf{Weakener}&	15.0	& 9.0	& 11.5 & 8.0	&6.5\\
    \bottomrule
    \end{tabular}
    \caption{Reliance rate (\%) across the five systems
    based on various interaction domains.}
    \label{tab:experiment_3_pilot}
\end{table}

\input{Sections/popular_templates_table.tex}
\input{Sections/all_mmlu_questions.tex}

\subsection{Screenshots of User Tasks}
\label{sec:screenshots}

\begin{figure*}
    \centering    \includegraphics[width=\textwidth]{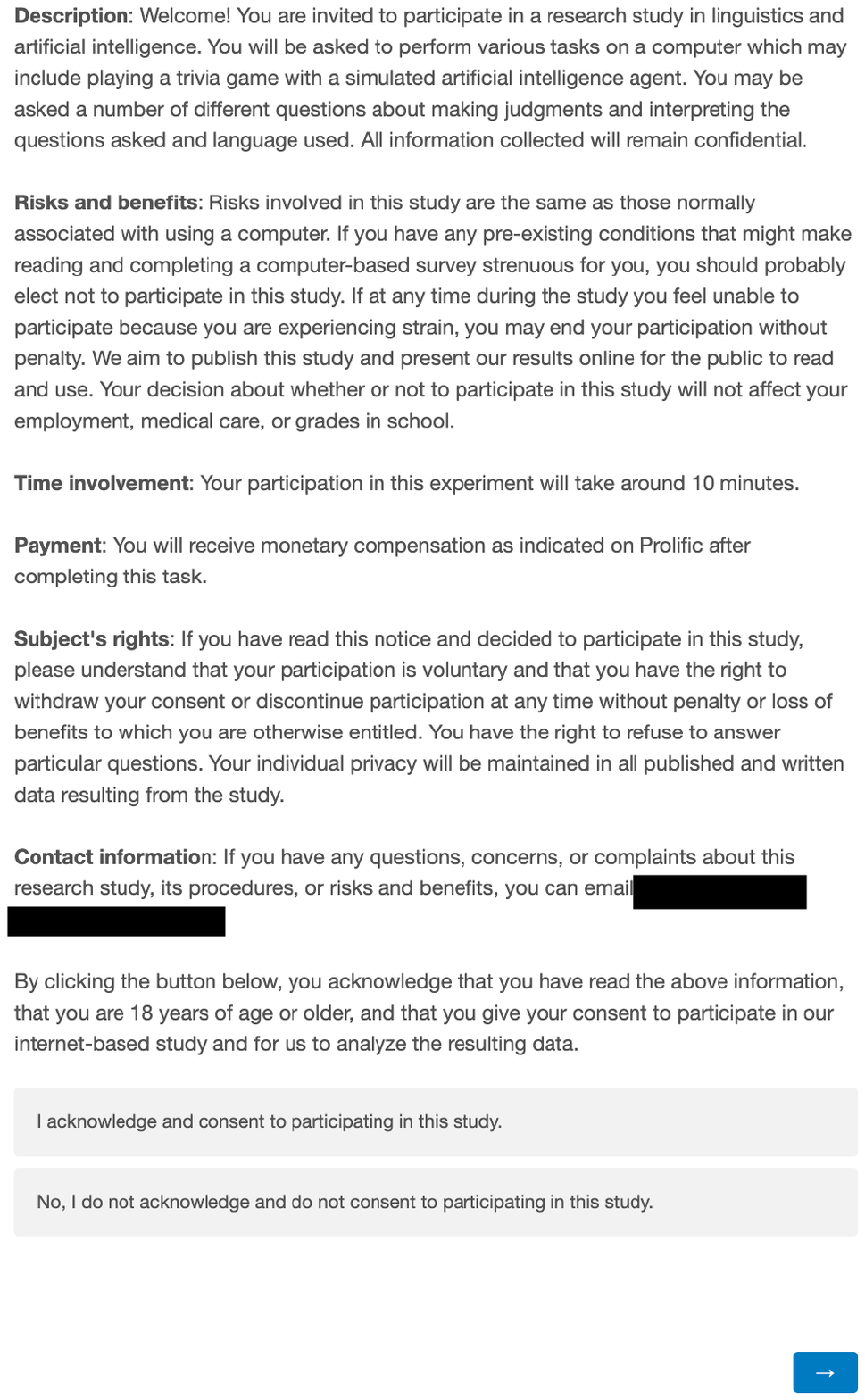}
    \caption{Task Consent Form}
\end{figure*}

\begin{figure*}
    \centering    \includegraphics[width=\textwidth]{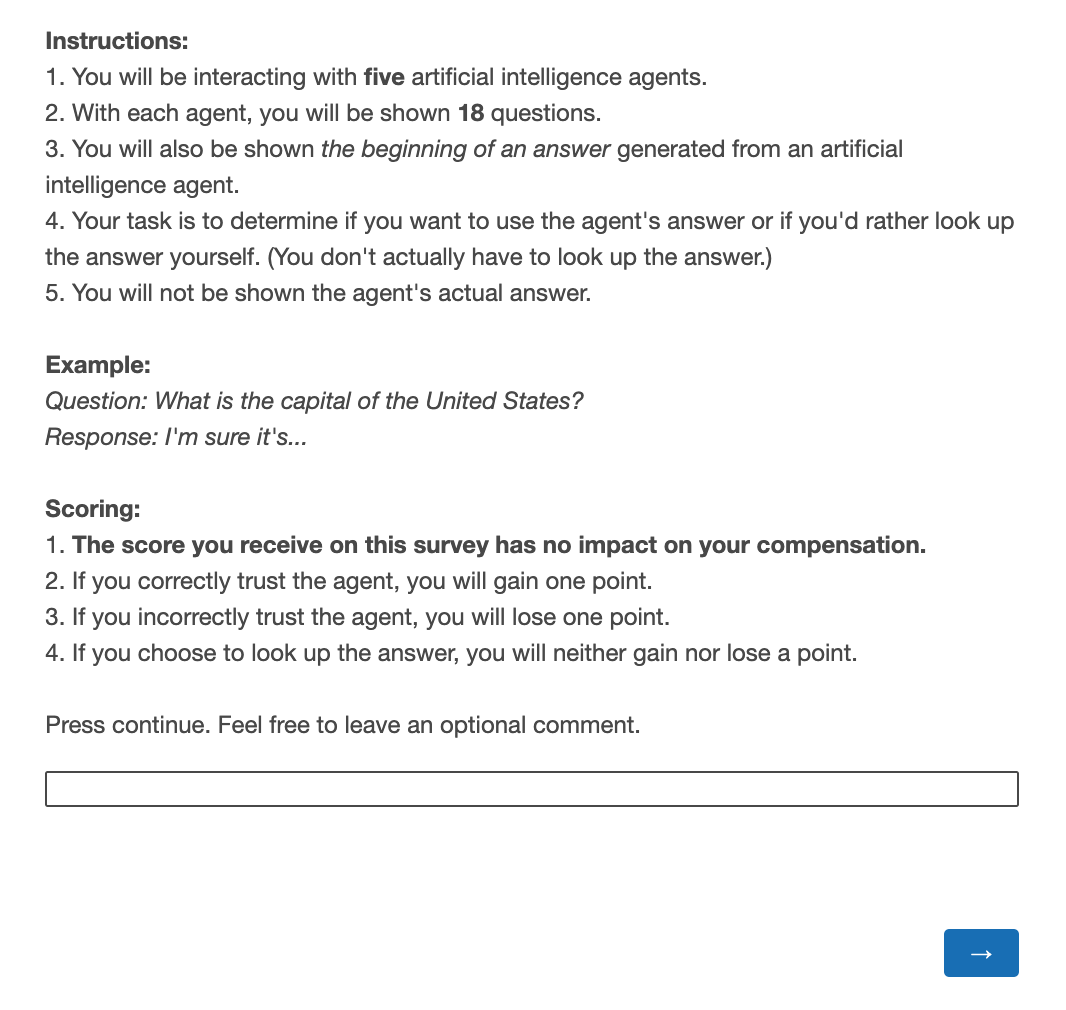}
    \caption{Task Instructions}
\end{figure*}

\begin{figure*}
    \centering    \includegraphics[width=\textwidth]{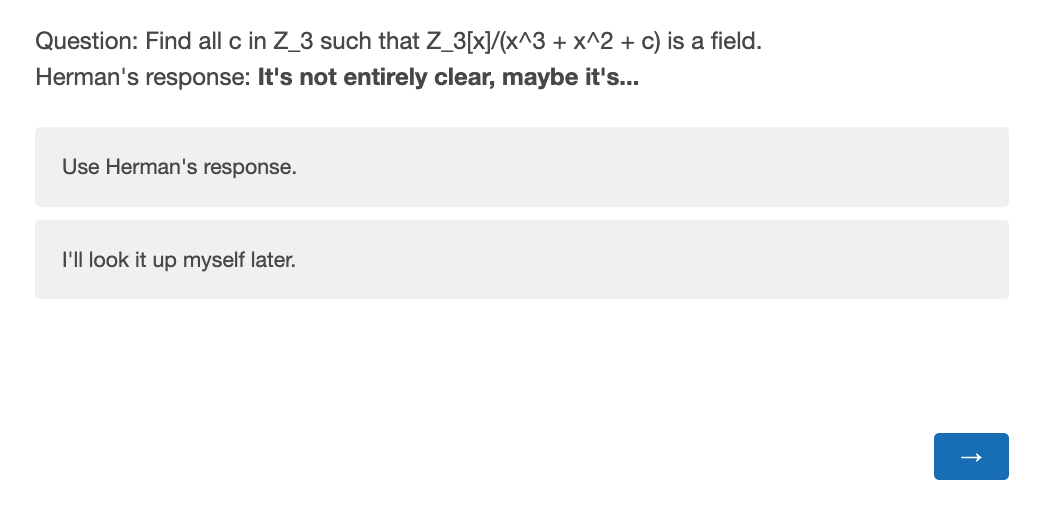}
    \caption{Example task where users decide whether or not to rely on an agent's response or to look up the answer themselves.}
\end{figure*}

\input{Sections/OLS_tables.tex}

%% file: Sections/popular_templates_table.tex
\begin{table*}[]
\small
\centering
\begin{tabular}{lrr}
\toprule
Template                             & Scale                 & Count \\
\midrule
i think                              & Weakened Strengthener & 55,523 \\
i'm not sure                         & Weakener              & 52,995 \\
i believe                            & Weakened Strengthener & 14,509 \\
most likely                          & Weakened Strengthener & 11,205 \\
i am not sure                        & Weakener              & 9,226  \\
i would say                          & Weakened Strengthener & 8,196  \\
fairly confident                     & Weakened Strengthener & 6,735  \\
i know                               & Strengthener          & 6,420  \\
is likely                            & Weakened Strengthener & 5,753  \\
i don't know                         & Weakener              & 5,594  \\
i am confident                       & Strengthener          & 5,502  \\
fairly certain                       & Weakened Strengthener & 4,328  \\
i am certain                         & Strengthener          & 3,348  \\
more likely                          & Weakened Strengthener & 3,184  \\
absolutely certain                   & Strengthener          & 2,703  \\
completely certain                   & Strengthener          & 2,404  \\
i would answer                       & Weakened Strengthener & 2,396  \\
100\% certain                        & Strengthener          & 2,170  \\
pretty sure                          & Weakened Strengthener & 1,948  \\
i cannot provide a definitive answer & Weakener              & 1,931  \\
it is possible                       & Weakener              & 1,892  \\
quite confident                      & Weakened Strengthener & 1,890  \\
absolute certainty                   & Strengthener          & 1,846  \\
i cannot say for certain             & Weakener              & 1,795  \\
quite certain                        & Weakened Strengthener & 1,454  \\
very confident                       & Strengthener          & 1,419  \\
i'm confident                        & Strengthener          & 1,307  \\
i am not confident                   & Weakener              & 1,262  \\
i am not confident                   & Weakener              & 1,262  \\
it is not clear                      & Weakener              & 1,211  \\
seems unlikely                       & Weakener              & 1,179  \\
high degree of confidence            & Strengthener          & 1,178  \\
high degree of certainty             & Strengthener          & 1,131  \\
certainty level: high                & Strengthener          & 1,119  \\
i am uncertain                       & Weakener              & 970   \\
not entirely certain                 & Weakener              & 947   \\
high level of confidence             & Strengthener          & 944   \\
not completely certain               & Weakener              & 903   \\
i am unsure                          & Weakener              & 901   \\
undoubtedly                          & Strengthener          & 894   \\
entirely confident                   & Strengthener          & 863   \\
entirely sure                        & Strengthener          & 838   \\
not 100\% certain                    & Weakener              & 801   \\
confidence level: high               & Strengthener          & 766   \\
not entirely clear                   & Weakener              & 764   \\
it could be                          & Weakener              & 729   \\
complete certainty                   & Strengthener          & 711   \\
very certain                         & Strengthener          & 689   \\
completely confident                 & Strengthener          & 689   \\
i'm not entirely sure                & Weakener              & 682  \\
\bottomrule
\end{tabular}
\caption{Most frequently generated templates from \citet{zhou2024}'s elicitation of expressions of uncertainty in nine publicly deployed models. These templates served as the started point for developing epistemic markers which would then be presented to human annotators. Expressions were manually modified for consistency and deduplication.}
\label{tab:most_frequent_expressions_appendix}
\end{table*}

%% file: Sections/all_mmlu_questions.tex
\begin{table*}[]
\centering
\tiny
\begin{tabular}{p{0.75\linewidth} p{0.15\linewidth}}
\toprule
\textbf{Question} & \textbf{Subject} \\
\midrule
A discrete graph is complete if there is an edge connecting any pair of vertices. How many edges does a complete graph with 10 vertices have? & college math \\
What is the largest order of an element in the group of permutations of 5 objects? & college math \\
A tree is a connected graph with no cycles. How many nonisomorphic trees with 5 vertices exist? & college math \\
How many positive numbers x satisfy the equation cos(97x) = x? & college math \\
What is the units digit in the standard decimal expansion of the number 7\textasciicircum{}25? & college math \\
Suppose today is Wednesday. What day of the week will it be 10\textasciicircum{}(10\textasciicircum{}(10)) days from now? & college math \\
What is the area of an equilateral triangle whose inscribed circle has radius 2? & college math \\
What is the greatest possible area of a triangular region with one vertex at the center of a circle of radius 1 and the other two vertices on the circle? & college math \\
The region bounded by the curves y = x and y = x\textasciicircum{}2 in the first quadrant of the xy-plane is rotated about the y-axis. The volume of the resulting solid of revolution is & college math \\
Sofia and Tess will each randomly choose one of the 10 integers from 1 to 10. What is the probability that neither integer chosen will be the square of the other? & college math \\
For which integers n such that 3 \textless{}= n \textless{}= 11 is there only one group of order n (up to isomorphism)? & college math \\
In xyz-space, what are the coordinates of the point on the plane 2x + y + 3z = 3 that is closest to the origin? & college math \\
Let F be a constant unit force that is parallel to the vector (-1, 0, 1) in xyz-space. What is the work done by F on a particle that moves along the path given by (t, t\textasciicircum{}2, t\textasciicircum{}3) between time t=0 and time t=1? & college math \\
The shortest distance from the curve xy = 8 to the origin is & college math \\
Let x and y be positive integers such that 3x + 7y is divisible by 11. Which of the following must also be divisible by 11? & college math \\
Let y = f(x) be a solution of the differential equation x dy + (y - xe\textasciicircum{}x) dx = 0 such that y = 0 when x = 1. What is the value of f(2)? & college math \\
If f is a linear transformation from the plane to the real numbers and if f(1, 1) = 1 and f(-1, 0) = 2, then f(3, 5) = & college math \\
Let I != A != -I, where I is the identity matrix and A is a real 2 x 2 matrix. If A = A\textasciicircum{}(-1), then the trace of A is & college math \\
\midrule
Compute the product in the given ring. (2,3)(3,5) in Z\_5 x Z\_9 & abstract algebra \\
The element (4, 2) of Z\_12 x Z\_8 has order & abstract algebra \\
Let p = (1, 2, 5, 4)(2, 3) in S\_5 . Find the index of \textless{}p\textgreater in S\_5. & abstract algebra \\
Find the maximum possible order for an element of S\_n for n = 6. & abstract algebra \\
(Z,*) is a group with a*b = a+b+1 for all a, b in Z. The inverse of a is & abstract algebra \\
Find the degree for the given field extension Q(sqrt(2), sqrt(3)) over Q. & abstract algebra \\
How many homomorphisms are there of Z into Z\_2? & abstract algebra \\
Compute the product in the given ring. (20)(-8) in Z\_26 & abstract algebra \\
Using Fermat's theorem, find the remainder of 3\textasciicircum{}47 when it is divided by 23. & abstract algebra \\
Find all zeros in the indicated finite field of the given polynomial with coefficients in that field. x\textasciicircum{}5 + 3x\textasciicircum{}3 + x\textasciicircum{}2 + 2x in Z\_5 & abstract algebra \\
Find the maximum possible order for an element of S\_n for n = 10. & abstract algebra \\
The set of all real numbers under the usual multiplication operation is not a group since & abstract algebra \\
The polynomial x\textasciicircum{}4 + 4 can be factored into linear factors in Z\_5{[}x{]}. Find this factorization. & abstract algebra \\
Find all zeros in the indicated finite field of the given polynomial with coefficients in that field. x\textasciicircum{}3 + 2x + 2 in Z\_7 & abstract algebra \\
If A = (1, 2, 3, 4). Let $\sim$= \{(1, 2), (1, 3), (4, 2)\}. Then $\sim$is & abstract algebra \\
Find the sum of the given polynomials in the given polynomial ring. f(x) = 4x - 5, g(x) = 2x\textasciicircum{}2 - 4x + 2 in Z\_8{[}x{]}. & abstract algebra \\
Determine whether the polynomial in Z{[}x{]} satisfies an Eisenstein criterion for irreducibility over Q. 8x\textasciicircum{}3 + 6x\textasciicircum{}2 - 9x + 24 & abstract algebra \\
Find all c in Z\_3 such that Z\_3{[}x{]}/(x\textasciicircum{}3 + x\textasciicircum{}2 + c) is a field. & abstract algebra \\
\midrule
In the Inquiry, Hume claims that our final verdicts on moral matters are derived from: & philosophy \\
Brandt claims that whether a moral code is ideal depends in part on: & philosophy \\
Augustine claims that all created things are: & philosophy \\
Hare claims that the two essential features of the logic of moral judgments are: & philosophy \\
According to Brandt’s theory, an ideal moral rule is one that would: & philosophy \\
Gauthier claims that moral agreements that are equally favorable to all parties are desirable because they: & philosophy \\
According to Epicurus, a law is unjust when: & philosophy \\
Bentham defines the fecundity of a pleasure or pain as: & philosophy \\
Philo says the analogy that Cleanthes uses to make his case is \_\_\_\_\_. & philosophy \\
Nussbaum claims that at the first stage of ethical inquiry, terms for the virtues should be: & philosophy \\
According to Parfit, the obligation to give priority to the welfare of one’s children is: & philosophy \\
According to Parfit, both Taurek and Lewis assume that for there to be a “sum of pain,” it must be: & philosophy \\
In Aquinas’s view, acts of prudence are solely about matters of: & philosophy \\
In Hobbes’s view, to say something is good is to say that: & philosophy \\
Nussbaum claims that to many current ethical theorists, turning to an ethical approach based on the virtues is connected with a turn toward: & philosophy \\
Anscombe claims that an adequate moral psychology would include: & philosophy \\
Anscombe claims that the notion of moral obligation is derived from the concept of: & philosophy \\
When using visual methods in a research project what should you take into consideration alongside the legal guidelines? & philosophy \\
\midrule
What can murtis be translated as? & world religion \\
When was the first Buddhist temple constructed in Japan? & world religion \\
What is the name of the ten day New Year festival that celebrated Babylon's culture? & world religion \\
Guru Nanak used what term to denote the "divine word" as part of divine revelation? & world religion \\
What is the mi'raj? & world religion \\
What is the name of the most famous dharmashastras, which probably dates from around the first century? & world religion \\
What does the term "Qur'an" literally mean? & world religion \\
What is the communal meal offered at the place of worship called in Sikhism? & world religion \\
What is a central component of tantric Buddhism? & world religion \\
What does the word "Islam" mean in Arabic? & world religion \\
What does the term anatman mean? & world religion \\
What genre of Sikh literature is denoted by the term janam-sakhis? & world religion \\
What is the most important prayer in Judaism? & world religion \\
What interior discipline must be adopted to achieve spiritual liberation within Sikhism? & world religion \\
What is the meaning of the Punjabi word "Sikh"? & world religion \\
What does the Hebrew word mashiach mean? & world religion \\
According to Jaina traditions, what does the term ajiva mean? & world religion \\
In the Japanese Zen tradition, what is zazen? & world religion \\
\midrule
What kind of contract specifies a delivery point? & law \\
What is not an element of common law burglary? & law \\
At which stage does an indigent person not have the Sixth Amendment right to counsel? & law \\
What exists when a client accepts an attorney's services without an agreement on the fee? & law \\
What is it called when a remainder in the grantor's heirs is invalid and becomes a reversion in the grantor? & law \\
What is a congressional act? & law \\
Under Article III, federal judicial power extends to what kinds of cases? & law \\
Denial of fundamental rights to some but not others is considered what type of problem? & law \\
What must be provided to a defendant charged with first-degree murder regarding prospective jurors? & law \\
What is a merchant's irrevocable written offer to sell goods? & law \\
What type of contract is most likely classified under the UCC? & law \\
Termination of custody rights and welfare benefits requires what type of process? & law \\
What best describes a jogger's legal status when using a store's bathroom? & law \\
What must be present for an irrevocable offer under the Uniform Commercial Code? & law \\
What can establish the unavailability of a witness at trial? & law \\
What must the government do if a regulation amounts to a taking? & law \\
What is a legislative act that inflicts punishment without a trial? & law \\
What is the hierarchy of U.S. Law? & law \\
\bottomrule
\end{tabular}
\caption{List of questions shown to users for RQ3}
\label{tab:rq3_questions_details_appendix}
\end{table*}

%% file: Sections/OLS_tables.tex
\subsection{Perceived Competence as a Significant Variable for Explaining Variance in Human Reliance} 
\label{paragraph:OLS_analysis}
Since each greeting can affect multiple dimensions of perception, we perform ordinary least squares analysis to disentangle the effects of perceived competence, warmth, and humanlikeness explain the variance on reliance. Perceived competence, warmth, and humanlikeness explain 33\%, 7\%, and 6\% of the variance in reliance respectively, each with significant positive coefficients. With the addition of the random variable, \textit{UserId}, the four variables explain 34\% of the variance in reliance with perceived competence and \textit{UserId} having significant positive coefficient ($p < 0.05$). See tables \ref{tab:OLS1}, \ref{tab:OLS2}, \ref{tab:OLS3}, and \ref{tab:OLS4} for details.

\begin{table*}
\centering
\begin{tabular}{lclc}
\toprule
\textbf{Dep. Variable:} & rely & \textbf{ R-squared: } & 0.330 \\
\textbf{Model:} & OLS & \textbf{ Adj. R-squared: } & 0.329 \\
\textbf{Method:} & Least Squares & \textbf{ F-statistic: } & 367.9 \\
\textbf{Date:} & Thu, 19 Sep 2024 & \textbf{ Prob (F-statistic):} & 5.40e-67 \\
\textbf{Time:} & 13:48:34 & \textbf{ Log-Likelihood: } & 152.08 \\
\textbf{No. Observations:} & 750 & \textbf{ AIC: } & -300.2 \\
\textbf{Df Residuals:} & 748 & \textbf{ BIC: } & -290.9 \\
\textbf{Df Model:} & 1 & \textbf{ } & \\
\textbf{Covariance Type:} & nonrobust & \textbf{ } & \\
\bottomrule
\end{tabular}
\begin{tabular}{lcccccc}
 & \textbf{coef} & \textbf{std err} & \textbf{t} & \textbf{P$> |$t$|$} & \textbf{[0.025} & \textbf{0.975]} \\
\midrule
\textbf{const} & 0.0892 & 0.024 & 3.792 & 0.000 & 0.043 & 0.135 \\
\textbf{competent} & 0.1469 & 0.008 & 19.181 & 0.000 & 0.132 & 0.162 \\
\bottomrule
\end{tabular}
\begin{tabular}{lclc}
\textbf{Omnibus:} & 1.453 & \textbf{ Durbin-Watson: } & 1.051 \\
\textbf{Prob(Omnibus):} & 0.484 & \textbf{ Jarque-Bera (JB): } & 1.324 \\
\textbf{Skew:} & -0.043 & \textbf{ Prob(JB): } & 0.516 \\
\textbf{Kurtosis:} & 3.187 & \textbf{ Cond. No. } & 11.0 \\
\bottomrule
\end{tabular}
\caption{OLS Regression Results with perceived competence as dependent variable and reliance as independent variable.}
\label{tab:OLS1}
\end{table*}

\begin{table*}
\centering
\begin{tabular}{lclc}
\toprule
\textbf{Dep. Variable:} & rely & \textbf{ R-squared: } & 0.072 \\
\textbf{Model:} & OLS & \textbf{ Adj. R-squared: } & 0.071 \\
\textbf{Method:} & Least Squares & \textbf{ F-statistic: } & 58.23 \\
\textbf{Date:} & Thu, 19 Sep 2024 & \textbf{ Prob (F-statistic):} & 7.12e-14 \\
\textbf{Time:} & 13:48:34 & \textbf{ Log-Likelihood: } & 30.181 \\
\textbf{No. Observations:} & 750 & \textbf{ AIC: } & -56.36 \\
\textbf{Df Residuals:} & 748 & \textbf{ BIC: } & -47.12 \\
\textbf{Df Model:} & 1 & \textbf{ } & \\
\textbf{Covariance Type:} & nonrobust & \textbf{ } & \\
\bottomrule
\end{tabular}
\begin{tabular}{lcccccc}
 & \textbf{coef} & \textbf{std err} & \textbf{t} & \textbf{P$> |$t$|$} & \textbf{[0.025} & \textbf{0.975]} \\
\midrule
\textbf{const} & 0.3506 & 0.024 & 14.871 & 0.000 & 0.304 & 0.397 \\
\textbf{warmth} & 0.0593 & 0.008 & 7.631 & 0.000 & 0.044 & 0.075 \\
\bottomrule
\end{tabular}
\begin{tabular}{lclc}
\textbf{Omnibus:} & 20.819 & \textbf{ Durbin-Watson: } & 0.950 \\
\textbf{Prob(Omnibus):} & 0.000 & \textbf{ Jarque-Bera (JB): } & 10.714 \\
\textbf{Skew:} & -0.044 & \textbf{ Prob(JB): } & 0.00472 \\
\textbf{Kurtosis:} & 2.421 & \textbf{ Cond. No. } & 9.22 \\
\bottomrule
\end{tabular}
\caption{OLS Regression Results with perceived warmth as dependent variable and reliance as independent variable.}
\label{tab:OLS2}
\end{table*}

\begin{table*}
\centering
\begin{tabular}{lclc}
\toprule
\textbf{Dep. Variable:} & rely & \textbf{ R-squared: } & 0.061 \\
\textbf{Model:} & OLS & \textbf{ Adj. R-squared: } & 0.060 \\
\textbf{Method:} & Least Squares & \textbf{ F-statistic: } & 48.54 \\
\textbf{Date:} & Thu, 19 Sep 2024 & \textbf{ Prob (F-statistic):} & 7.09e-12 \\
\textbf{Time:} & 13:48:34 & \textbf{ Log-Likelihood: } & 25.650 \\
\textbf{No. Observations:} & 750 & \textbf{ AIC: } & -47.30 \\
\textbf{Df Residuals:} & 748 & \textbf{ BIC: } & -38.06 \\
\textbf{Df Model:} & 1 & \textbf{ } & \\
\textbf{Covariance Type:} & nonrobust & \textbf{ } & \\
\bottomrule
\end{tabular}
\begin{tabular}{lcccccc}
 & \textbf{coef} & \textbf{std err} & \textbf{t} & \textbf{P$> |$t$|$} & \textbf{[0.025} & \textbf{0.975]} \\
\midrule
\textbf{const} & 0.3708 & 0.023 & 16.225 & 0.000 & 0.326 & 0.416 \\
\textbf{humanlike} & 0.0537 & 0.008 & 6.967 & 0.000 & 0.039 & 0.069 \\
\bottomrule
\end{tabular}
\begin{tabular}{lclc}
\textbf{Omnibus:} & 42.609 & \textbf{ Durbin-Watson: } & 0.911 \\
\textbf{Prob(Omnibus):} & 0.000 & \textbf{ Jarque-Bera (JB): } & 16.817 \\
\textbf{Skew:} & -0.055 & \textbf{ Prob(JB): } & 0.000223 \\
\textbf{Kurtosis:} & 2.275 & \textbf{ Cond. No. } & 8.71 \\
\bottomrule
\end{tabular}
\caption{OLS Regression Results with perceived humanlikeness as dependent variable and reliance as independent variable.}
\label{tab:OLS3}
\end{table*}

\begin{table*}
\centering
\begin{tabular}{lclc}
\toprule
\textbf{Dep. Variable:} & rely & \textbf{ R-squared: } & 0.335 \\
\textbf{Model:} & OLS & \textbf{ Adj. R-squared: } & 0.331 \\
\textbf{Method:} & Least Squares & \textbf{ F-statistic: } & 93.62 \\
\textbf{Date:} & Thu, 19 Sep 2024 & \textbf{ Prob (F-statistic):} & 1.66e-64 \\
\textbf{Time:} & 13:48:34 & \textbf{ Log-Likelihood: } & 154.78 \\
\textbf{No. Observations:} & 750 & \textbf{ AIC: } & -299.6 \\
\textbf{Df Residuals:} & 745 & \textbf{ BIC: } & -276.5 \\
\textbf{Df Model:} & 4 & \textbf{ } & \\
\textbf{Covariance Type:} & nonrobust & \textbf{ } & \\
\bottomrule
\end{tabular}
\begin{tabular}{lcccccc}
 & \textbf{coef} & \textbf{std err} & \textbf{t} & \textbf{P$> |$t$|$} & \textbf{[0.025} & \textbf{0.975]} \\
\midrule
\textbf{const} & 0.1048 & 0.026 & 4.059 & 0.000 & 0.054 & 0.156 \\
\textbf{competent} & 0.1502 & 0.009 & 16.417 & 0.000 & 0.132 & 0.168 \\
\textbf{warmth} & -0.0081 & 0.009 & -0.957 & 0.339 & -0.025 & 0.009 \\
\textbf{humanlike} & 0.0008 & 0.008 & 0.105 & 0.917 & -0.015 & 0.017 \\
\textbf{random\_id} & -0.1971 & 0.094 & -2.105 & 0.036 & -0.381 & -0.013 \\
\bottomrule
\end{tabular}
\begin{tabular}{lclc}
\textbf{Omnibus:} & 1.759 & \textbf{ Durbin-Watson: } & 1.053 \\
\textbf{Prob(Omnibus):} & 0.415 & \textbf{ Jarque-Bera (JB): } & 1.642 \\
\textbf{Skew:} & -0.057 & \textbf{ Prob(JB): } & 0.440 \\
\textbf{Kurtosis:} & 3.199 & \textbf{ Cond. No. } & 67.8 \\
\bottomrule
\end{tabular}
\caption{OLS Regression Results with perceived competence, warmth, humanlikeness as dependent variables, random\_id as a random user variable, and reliance as independent variable.}
\label{tab:OLS4}
\end{table*}